\documentclass[10pt,twocolumn]{article}

\usepackage[utf8]{inputenc}
\usepackage[T1]{fontenc}
\usepackage{amsmath,amssymb,amsfonts}
\usepackage{graphicx}
\usepackage{subcaption}
\usepackage{booktabs}
\usepackage{caption}
\usepackage[margin=0.75in,columnsep=0.25in]{geometry}
\usepackage{hyperref}
\usepackage{authblk}
\usepackage{algorithm}
\usepackage{algorithmic}
\usepackage[numbers,sort&compress]{natbib}
\usepackage{siunitx}
\usepackage{makecell}

\title{Infrastructure-Based Multi-Agent Reinforcement Learning with Attention for Highway Ramp Merging}
\author[1]{Chenhao Tong}
\author[1]{Steffen Müller}
\affil[1]{Technische Universität Berlin, Berlin, Germany\\
\texttt{\{chenhao.tong, steffen.mueller\}@tu-berlin.de}}
\date{}

\begin{document}
\twocolumn[
  \begin{@twocolumnfalse}
    \maketitle
    \begin{abstract}
    
    On-ramp merging is a typical safety and throughput problem on highways. Without proper coordination, human drivers and self-interested autonomous vehicles may make risky manoeuvres and cause accidents or congestion. With recent advances in connected autonomous vehicles (CAVs) and multi-agent reinforcement learning (MARL), vehicles can be trained to coordinate for safe and efficient merging. However, in dense traffic, forming an effective observation of the traffic state is challenging. For example, observing only the neighbouring vehicles may lead to locally optimal decisions, while observing the entire traffic can introduce much irrelevant information. In this paper, we propose a transformer-based MARL vehicle control model to address these problems. A roadside infrastructure controller receives the states of all CAVs via vehicle-to-infrastructure (V2I) communication and sends control commands back to them. An attention-based observation encoder is designed where each vehicle can filter the relevant vehicle information from the global traffic state. The proposed model is evaluated in SUMO against three baselines: a model that observes only the neighbouring vehicles, a model that encodes the global traffic information, and the rule-based SUMO model. The results show that our model achieves the highest traffic flow, safety and smoothness. Furthermore, in a four-hour simulation with dynamic traffic demand, our model successfully avoids traffic jams, while other baseline models fail at high traffic flow. We also show that our model remains effective when human-driven vehicles are present. We further show that our model, trained in ideal communication conditions, can still remain collision-free and maintain throughput under realistic V2I communication errors. Finally, we visualise the relevance weights that an ego vehicle assigns to other vehicles in multiple traffic situations.
    
    \end{abstract}
    \vspace{1em}
  \end{@twocolumnfalse}
]


\section{Introduction}

Highway ramp merging is a typical safety and traffic flow control problem on highways. At a highway merge point, two traffic streams with high density and high speed must converge into one shared lane within a limited distance. Because of the intense interaction between vehicles, systematic traffic coordination is essential for safe and efficient merging. Despite the highway merging rules, human drivers follow those rules inconsistently and may make risky manoeuvres. For example, drivers on the ramp may squeeze into gaps without maintaining a safe distance, drivers on the main road may speed up and refuse to yield, or drivers may hesitate when it is unclear who should go first \cite{ramp-driver-beh}. As urbanisation accelerates, the increasing number of private vehicles and autonomous vehicles worsens the problem. Without coordination, drivers and self-interested driving automation systems can optimise only their own routes and speeds, without accounting for one another. This further intensifies these conflicts. As a result, a solution that systematically coordinates vehicles for safe and efficient merging behaviour is urgently needed.

Connected and autonomous vehicles (CAVs) \cite{cav-survey} offer a promising framework to address the merging problem. In a CAV system, vehicles can communicate to share information and coordinate their behaviours instead of each making an individual judgement. Recent progress in artificial intelligence has further extended the ability of CAVs from following pre-defined driving rules to perceiving and analysing their surroundings and making driving decisions accordingly. Among these techniques, multi-agent reinforcement learning (MARL) \cite{mappo} is particularly well suited to this problem, as it allows multiple CAVs to learn cooperative merging behaviour through interaction with one another in the traffic environment. This multi-agent learning-based approach has been shown to achieve state-of-the-art performance in many traffic coordination problems, such as traffic signal control \cite{traffic-light}, intersection coordination \cite{intersection}, and traffic platooning \cite{platoon}. This success has motivated the extension of MARL methods to the ramp merging problem \cite{ramp-survey}, which has achieved state-of-the-art performance under high-density and dynamic traffic conditions \cite{CAE, ramp-marl-performance, GNN1, GNN2}.

In MARL approaches, each CAV needs to form an observation of its surrounding environment, e.g., other vehicles' speed and position, for further decision-making. To obtain this information, a typical approach is to use the vehicle-to-vehicle (V2V) communication framework, where CAVs broadcast their local states to other vehicles to obtain a full observation of the control area \cite{CAE}. However, a limitation of this framework is message overhead: if every vehicle communicates with every other vehicle, the number of message exchanges grows quadratically with the number of communicating vehicles. In dense traffic, real-time communication among hundreds of vehicles becomes impractical. In addition, these approaches typically assume ideal communication conditions, whereas a high communication channel load increases message loss and delay, and intensive message processing adds computational delay. As a result, vehicles may act based on incomplete or outdated information. More critically, even if such communication were achieved, filtering the relevant information from so many vehicle states remains a significant challenge. One line of work reduces the communication overhead by restricting each vehicle to communicate only with its neighbours \cite{CTDE-baseline}. However, this also limits vehicles' awareness of traffic conditions elsewhere in the control area, which may lead to sub-optimal performance.

With recent advances in large language models (LLMs), the self-attention neural network architecture has been used to encode large contexts by filtering for only the relevant information with respect to each word \cite{attention}. Similarly, in dense traffic, each CAV can use this mechanism to filter for the information of other vehicles that is relevant to its merging decision. In this work, we aim to improve existing MARL-based approaches to highway ramp merging by using the self-attention mechanism to encode relevant traffic observations from the whole control area for every vehicle without excessive communication overhead. Our contributions are listed as follows:

\begin{enumerate}
    \item We propose a hierarchical MARL framework relying on vehicle-to-infrastructure (V2I) communication for minimal communication overhead. CAVs report their local states to a roadside control unit, which then issues high-level commands to each vehicle to execute. Therefore, each execution loop contains only one message round trip.

    \item We propose an attention-based architecture that learns which other vehicles in the control area are most relevant to each vehicle's merging decision. This allows each vehicle to form a global observation that retains only the relevant information for efficient control. We further visualise each vehicle's attention weights to understand which traffic regions it focuses on in different traffic situations.

    \item We compare our proposed model against three baselines: a neighbour observation-based MARL method, a global observation-based MARL method, and a rule-based merging strategy. The results show that our model achieves the best performance in safety, traffic throughput and smoothness across multiple experimental setups.

    \item We further evaluate the model under mixed traffic conditions with human-driven vehicle (HDV) penetration rates ranging from 10\% to 40\%. The proposed model achieves varying degrees of traffic performance optimisation across these penetration rates.

    \item We further evaluate the model performance under realistic V2I communication constraints  -- message loss, delay and jitter -- and vehicle sensor noise. Without retraining, the proposed model remains collision-free and maintains traffic throughput.
    
\end{enumerate}

The remainder of the paper is organised as follows. In Section 2, we review the related work on MARL-based on-ramp merging approaches. In Section 3, we present the proposed hierarchical MARL framework, including the V2I communication scheme, the observation and action spaces, the reward function, the attention-based architecture, and the training algorithm. In Section 4, we describe the experiment setup and present the results. We conclude the paper and discuss directions for future work in Section 5.

\section{Related Work}

Prior MARL-based highway ramp merging approaches largely model the merging problem as a single-vehicle problem, where an autonomous vehicle observes its surrounding vehicles and learns the optimal behaviour for merging accordingly \cite{single1, single2}. However, when multiple CAVs are present, their independently optimised decisions can conflict, e.g., several vehicles want to merge into the same gap at the same time regardless of one another. This may degrade the traffic throughput and safety at the merge point \cite{non-coop}. Therefore, recent merging approaches model the merging problem as a multi-agent problem, where CAVs learn to coordinate jointly rather than act independently. 

Existing MARL-based solutions commonly model the multi-agent CAV merging scenario as a bounded control area surrounding the merge point \cite{ramp-marl-performance, CAE, GNN1, GNN2, CTDE-baseline, sin-curve, csac-merging}. Within the control area, vehicles can observe their own state and communicate with each other to share information. The merging objective is typically formulated as a multi-objective optimisation problem, considering safety and traffic throughput. Their control model architectures are typically structured hierarchically, in which a high-level reinforcement learning policy issues a coarse manoeuvre -- e.g. change to the left/right lane or speed up/slow down -- while a low-level controller executes the corresponding trajectory using techniques such as optimal control \cite{CAE}, a predefined kinematic trajectory model \cite{sin-curve}, a PID controller \cite{non-coop}, or model predictive control \cite{single1}.

However, these approaches differ in vehicles' communication protocols and how vehicles form observations. One line of work relies on full vehicle-to-vehicle (V2V) communication, where every CAV broadcasts its state to others \cite{GNN1, GNN2} to form a global observation of the control area traffic. However, if every vehicle communicates with each other, the resulting communication overhead grows quadratically with the number of vehicles in the control zone. Therefore, in dense traffic, achieving real-time communication between vehicles is challenging. In addition, in dense traffic, processing the resulting large joint vehicle states is also difficult. These methods typically encode the joint vehicle states using structured neural network architectures. For example, graph neural networks are used to aggregate information from vehicles connected to the ego vehicle under a predefined adjacency rule, such as a distance threshold \cite{GNN1, GNN2}, or the Convolutional Autoencoder (CAE) compresses the full traffic state into a fixed-size grid representation \cite{CAE}. In both cases, however, the relevance of one vehicle to another is determined by the neural architecture itself rather than learned from data. A convolutional filter in the CAE applies the same fixed spatial weights everywhere, and a graph's edge weights are determined by a hand-designed rule, e.g., depending on the distances between the vehicles. As a result, which vehicles matter to a merging decision-making is fixed in advance, rather than determined by the actual traffic situation, potentially leading to sub-optimal performance in high-density, complex merging scenarios.

Other lines of work restrict this exchange to each vehicle's immediate neighbours or a fixed number of vehicles within a pre-defined range. This approach effectively bounds the vehicle's observation size and thus reduces communication overhead as traffic density increases \cite{sin-curve, CTDE-baseline}. However, restricting each vehicle's observation to its immediate surroundings turns the decision problem into a partially observable one, where vehicles have no information about the global traffic. This potentially also results in suboptimal decisions \cite{POMDP}.

One way to avoid heavy communication overhead among vehicles while still allowing a global observation is to adopt a vehicle-to-infrastructure (V2I) framework: each vehicle reports its own state to a central roadside control unit, which then computes and transmits actions for each vehicle to execute. This leaves open how the control unit should efficiently process the resulting volume of vehicle information. Rather than relying on predefined communication partners, such as a fixed set of neighbouring vehicles, one approach is to let each vehicle learn who it should attend to based on the current traffic scene. A potential solution is the self-attention network from large language models, which learns relevance weights between words based on the text content \cite{attention}. In this work, we adapt the self-attention mechanism and develop a traffic information encoder where each vehicle learns to assign relevance weights to other vehicles based on their states. Therefore, it can focus on the vehicles that matter to its merging decision.

However, many prior works assume ideal vehicle and communication conditions. For example, vehicles can measure their state accurately through sensors, and they can communicate instantly without any errors~\cite{GNN1, GNN2, CAE, CTDE-baseline, sin-curve}. In reality, V2V and V2I links suffer from various communication errors, such as message loss, delay and jitter, and vehicle states contain sensor noise. The impact of these real-world constraints on the vehicle control policies remains largely unexplored. 

In the next section, we describe our attention-based network architecture and our multi-agent reinforcement learning algorithm.

\section{Method}
\label{sec:method}
In this section, we introduce how we formulate the highway ramp merging problem as a multi-agent reinforcement learning task. We describe the infrastructure-based multi-vehicle controller, define vehicles' observations, actions, and reward function design. We detail the proposed self-attention-based policy network architectures and the modified proximal policy optimisation (PPO)~\cite{ppo} training algorithm.
\begin{figure*}[t!]
    \centering
    \includegraphics[width=\textwidth]{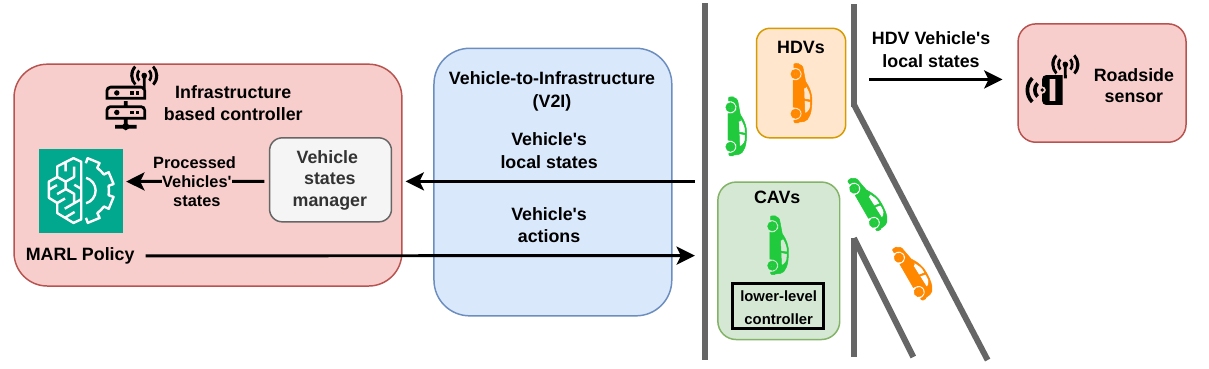}
    \caption{The hierarchical ramp merging system architecture. The infrastructure-based controller receives states from connected vehicles (CAVs) via the V2I network and from HDVs through roadside sensors. A vehicle states manager handles vehicles entering and leaving the controlled area and adds additional information to the received states. The MARL traffic control policy takes the processed states as input and issues actions to CAVs only, and the CAVs’ lower-level controllers execute these actions.}
    \label{fig:architecture}
\end{figure*}

\subsection{Problem Formation}
\label{subsec:framework}
Following the common approach in the literature \cite{CTDE-baseline, CAE, sin-curve, single1}, we use a hierarchical control architecture to coordinate vehicles at two levels. Our MARL traffic control policy governs tactical decision-making for each vehicle, i.e., it decides when the vehicle should accelerate, decelerate, or change lanes. CAVs then incorporate a lower-level controller, e.g., a PID or MPC, to execute the tactical decisions. 

Fig.~\ref{fig:architecture} illustrates our proposed system architecture. In this work, we use an infrastructure-based controller that hosts the MARL traffic control policy. CAVs report their local states to the controller over the V2I network and receive tactical actions from it. We assume that HDVs do not have hardware for V2I communication and their states are instead perceived through roadside sensors. A vehicle states manager then processes each vehicle’s state report, i.e., it manages vehicles entering and leaving, adds additional information to each vehicle’s state, and formats the states for the control policy. In addition, it assigns a unique ID to an entering vehicle, and it clears the vehicle ID and stored information when the vehicle exits the control region.

This centralised architecture contrasts with the decentralised architecture used in prior work \cite{CTDE-baseline}. Routing all communication to a centralised infrastructure controller bounds the communication overhead to a single exchange per vehicle, keeping communication scalable as the number of vehicles grows.



In this work, we adopt three criteria to evaluate the local and global collaborative performance of highway ramp merging traffic control policies: safety, traffic throughput, and smoothness. These criteria reflect whether the control policy can maximise the overall traffic flow and resolve local conflicts between vehicles, while ensuring safety compliance and smoothness.


\subsection{Vehicle Observation Space}
\label{subsec:observation}

In our centralised architecture, each vehicle observes its own state and forwards it to the infrastructure-based controller. The observation of a vehicle consists of its position and kinematics. Kinematics information includes the vehicle's longitudinal and lateral speed, longitudinal and lateral acceleration, and its heading angle ($\psi$), which is encoded as $(\sin\psi, \cos\psi)$. A vehicle’s position is given by its current GPS coordinates.

After receiving each vehicle’s local states, the vehicle states manager augments them with the vehicle’s road position context, control context and type. The control context includes the current road speed limit, the remaining lane-change cooldown time (see Section~\ref{subsec:action}), and the vehicle’s previous action. Since HDVs are not controlled by the MARL policy, their control context fields are filled with zeros. A vehicle’s road position context is the coordinate of its current road segment in the road graph, given by the Laplacian positional encoding~\cite{Laplacian}. A vehicle’s type is a binary flag indicating whether it is a CAV or an HDV.


Vehicle $i$'s observation space vector is therefore:
\begin{equation}
    o_i = \left[\, x_i,\ y_i,\ \mathbf{k}_i,\ \mathbf{c}_i,\ \mathbf{p}_i,\ t_i \,\right],
\end{equation}
where $[x_i,\ y_i]$ is the relative vehicle position in the control region, calculated from its GPS coordinates by the vehicle states manager, $\mathbf{k}_i = [v^{\mathrm{lon}}_i,\ v^{\mathrm{lat}}_i,\ a^{\mathrm{lon}}_i,\ a^{\mathrm{lat}}_i,\ \sin\psi_i,\ \cos\psi_i]$ is the kinematics, $\mathbf{c}_i$ the control context, $\mathbf{p}_i$ the road position context, and $t_i \in \{0, 1\}$ the vehicle type.

\subsection{Action Space}
\label{subsec:action}
In each decision step, the proposed control policy selects one of five tactical actions for each CAV: \texttt{Cruise}, \texttt{Lane Change Left}, \texttt{Lane Change Right}, \texttt{Lower Speed Limit}, and \texttt{Raise Speed Limit}. The Cruise action maintains the vehicle’s current lane and speed limit. The two lane-change actions move the vehicle one lane to the left or right. After a lane change, the vehicle must wait for a cooldown period before it can perform another one. This allows the vehicle’s lower-level controller to complete the manoeuvre and avoids unrealistically frequent lane changes. Instead of directly adjusting the vehicle’s actual speed, the two speed-limit actions only adjust its speed limit by a fixed increment. The speed limit is bounded between zero and the road’s speed limit. The vehicle’s lower-level controller aims to drive at or near the set speed limit, but it can still slow down the vehicle when required for safety, for example, to keep a safe gap to a slower leading vehicle.

As an additional safety measure, we adopt the action masking method~\cite{starcraft}, which restricts the control policy to sampling only from currently safe actions during runtime, instead of only relying on the reward function, which discourages unsafe actions during training. For example, a lane-change action can be masked out if the vehicle's cooldown from a previous manoeuvre has not yet elapsed, the target lane has no sufficient gap, or the vehicle is driving in a junction where a lane change is not allowed. This guarantees a safe exploration process during training and helps the policy converge faster~\cite{starcraft}. The speed-limit actions are not masked, because raising the speed limit does not force the vehicle to speed up. For example, if the leading vehicle is too close and driving slowly, the lower-level vehicle controller will decelerate the vehicle to keep a safe gap, regardless of the set speed limit.

\subsection{Reward Function}
\label{subsec:reward}
The reward function evaluates each vehicle's performance in the merging scenario. In this work, we focus on three performance criteria: safety, traffic throughput, and traffic smoothness. As safety relates to a vehicle's local performance while traffic throughput and smoothness are outcomes of the entire traffic flow, each vehicle's reward is therefore a combination of a local term and a global term.

\paragraph{Local reward.}
The local reward measures the safety and progress of each vehicle’s own driving behaviour. For safety, a collision is penalised for the colliding vehicle, as a collision is mainly caused by the vehicle’s own manoeuvre decision, such as changing lanes without a safe gap or poor speed control. The higher the relative speed between the vehicles at the time of the collision, the higher the penalty. For progress, a reward is assigned to the vehicle according to its speed to encourage the vehicle to drive faster and potentially increase the traffic flow. However, the speed reward is only a small part of the local reward, because the globally optimal traffic flow requires collaboration between vehicles, such as giving way to others and changing to a less occupied lane in advance, instead of greedily maximising their own speed.

Eq.~\ref{eq:local_reward} shows the local reward $r^{\mathrm{loc}}_{i,t}$ of vehicle $i$ at step $t$.
\begin{equation}
r^{\mathrm{loc}}_{i,t} =
\begin{cases}
\dfrac{v_{i,t}}{\bar{v}_{i,t}} - (\kappa_0 + \kappa_s\, s_{i,t}) - \beta, & \text{if vehicle } i \text{ collides},\\[2ex]
\dfrac{v_{i,t}}{\bar{v}_{i,t}} - \beta, & \text{otherwise},
\end{cases}
\label{eq:local_reward}
\end{equation}
where $v_{i,t}$ is the speed of vehicle $i$ and $\bar{v}_{i,t}$ is the speed limit. $\kappa_0$ is the penalty assign to vehicle $i$ if it causes a collision, and $s_{i,t}$ is the collision severity, defined as the relative speed between vehicle $i$ and the collided vehicle at the moment of collision. The severity is scaled by $\kappa_s$. The collision penalty is applied only at the step in which the collision occurs, so a persisting collision state is only penalised once. The offset $\beta$ is a local reward scalar.

\paragraph{Global reward.}
The global reward evaluates the overall traffic collaboration performance, which results from the joint behaviour of all vehicles. In this work, we consider two criteria reflecting this performance: traffic flow and hard brake rate. The traffic flow measures the number of vehicles passing through the merging area per unit time. It reflects the shared interest of all vehicles. The hard brake rate measures the smoothness of the traffic, reflecting the conflict of interest between vehicles: a vehicle that does not consider other vehicles’ actions when planning its movements can force others to slow down or brake hard. 

Eq.~\ref{eq:global_reward} describes the global reward at time step $t$.

\begin{equation}
G_t = \lambda_{\mathrm{tp}}\big(T^{\mathrm{main}}_t + T^{\mathrm{merge}}_t\big) - \lambda_{\mathrm{eb}}\, n^{\mathrm{eb}}_t,
\label{eq:global_reward}
\end{equation}
where $T^{\mathrm{main}}_t$ and $T^{\mathrm{merge}}_t$ are the throughputs of the main road and the merge ramp at step $t$, i.e., the number of vehicles exiting the control region and the number of vehicles merging from the ramp onto the main road, respectively. $n^{\mathrm{eb}}_t$ is the number of hard brakes in step $t$, and $\lambda_{\mathrm{tp}}$ and $\lambda_{\mathrm{eb}}$ are scalars.

\paragraph{Combined reward.}
Each vehicle $i$ in the observation zone receives the combined reward $r_{i,t}$ at time step $t$.
\begin{equation}
r_{i,t} = (1-\alpha)\, r^{\mathrm{loc}}_{i,t} + \alpha\, G_t,
\label{eq:reward}
\end{equation}
where $\alpha \in [0,1]$ weights the global reward and the local reward. 
\subsection{Traffic Control Neural Network}
\label{subsec:architectures}
When a vehicle is making tactical decisions in the ramp merge problem, it should focus only on the vehicles that are relevant to its current state and intention. For example, an on-ramp vehicle should focus on the main road vehicles that it will interact with when it reaches the merge point, as well as its leading and following vehicles, to maintain a safe gap and smooth interaction.

To let the vehicle determine the relevance of every other vehicle to itself, we propose a transformer-based traffic control neural network. The control neural network contains three components: a vehicle observation encoder, which embeds a vehicle’s historical observation; a transformer encoder, which determines the relevance between vehicles; and the actor and critic networks, which output the action distributions and estimated state values.

\begin{figure}[t]
    \centering
    \includegraphics[width=\linewidth]{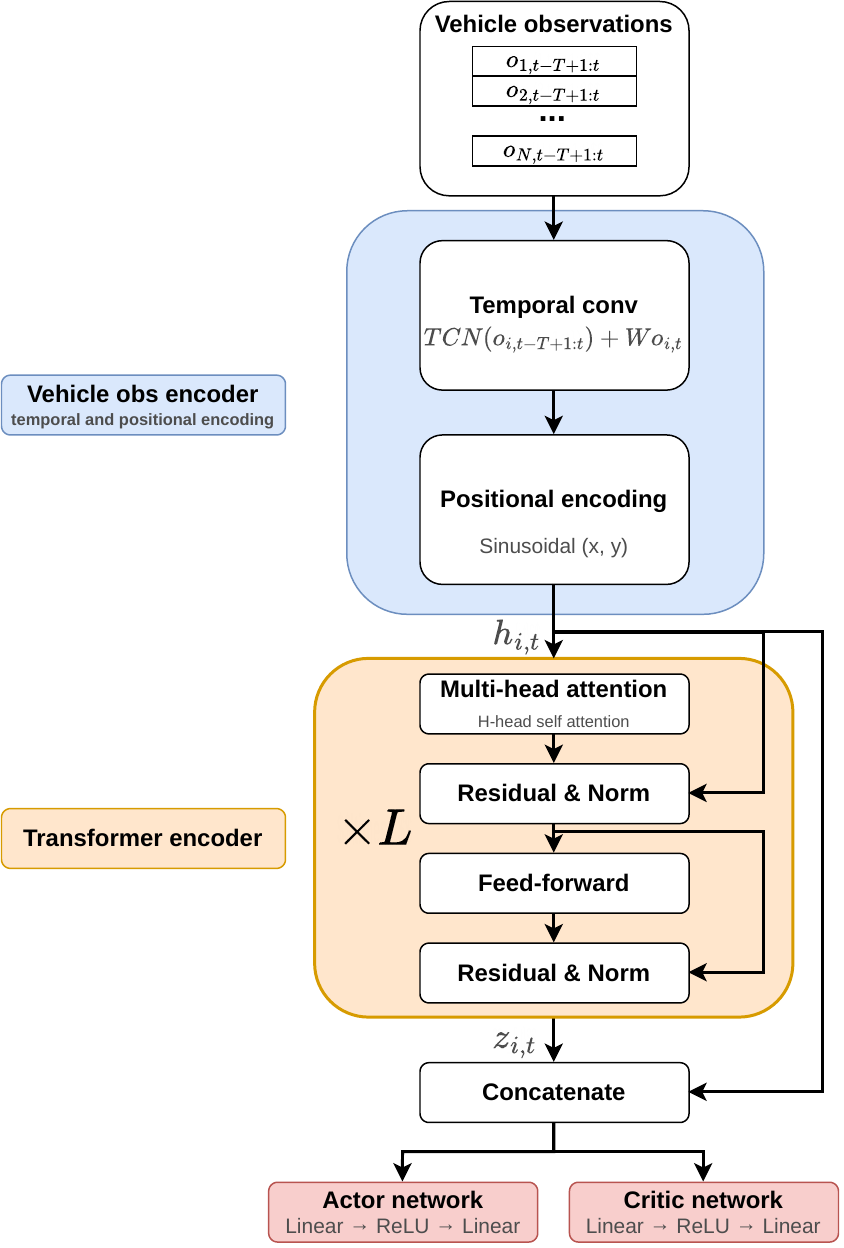}
    \caption{Architecture of the proposed network.}
    \label{fig:nn-architecture}
\end{figure}

The architecture of the control neural network is shown in Fig.~\ref{fig:nn-architecture}.

\paragraph{Vehicle observation encoding.}

A single vehicle observation at one timestep, defined in Section~\ref{subsec:observation}, does not encode a vehicle's recent manoeuvres or intentions, such as whether it has just started a lane change. Therefore, the vehicle states manager keeps an observation history of vehicle $i$, $\mathbf{o}_{i,t-T+1:t} \in \mathbb{R}^{T\times F}$, as input, where $T$ is the history length and $F$ is the dimension of the observation space. The vehicle observation encoder then embeds the observation history of vehicles. 

As vehicles only remain in the control region for a short period, a temporal convolutional network (TCN)~\cite{tcn} is used to encode $\mathbf{o}_{i,t-T+1:t}$, as it trains faster and more stably than recurrent networks for short history encoding. Vehicle $i$'s current state $\mathbf{W}\,\mathbf{o}_{i,t}$ is added to the TCN encoding through a residual connection, because a vehicle's current state is also important for decision making. Here, $\mathbf{W}$ is a learnable projection which matches the size of $\mathbf{o}_{i,t}$ to the size of the TCN output. 

Although a vehicle's observation contains its position in the control region, we use a sinusoidal encoding~\cite{attention} $\mathrm{PE}(\mathbf{p}_{i,t})$ to map the position $\mathbf{p}_{i,t}=(x,y)$ to multi-frequency features, which allow the attention to capture relative positions between vehicles more easily than raw coordinates. 

The vehicle observation encoding $\mathbf{h}_{i,t}$ of vehicle $i$ at timestep $t$ is given by
\begin{equation}
\mathbf{h}_{i,t} = \mathrm{TCN}\big(\mathbf{o}_{i,t-T+1:t}\big) + \mathbf{W}\,\mathbf{o}_{i,t} + \mathrm{PE}(\mathbf{p}_{i,t}).
\label{eq:vehicle_encoding}
\end{equation}

\paragraph{Attention over vehicles.}
The relevance of other vehicles to vehicle $i$ depends on their current state and their intentions. We use self-attention~\cite{attention} to determine the relevance between vehicles. In self-attention, each vehicle's state embedding plays three roles: as a query, which expresses what information the vehicle requires to decide an action; as a key, which is matched against the queries of other vehicles to determine its relevance to them; and as a value, which provides the state information to the vehicles that attend to it.

For vehicle $i$, the query, key and value are obtained by learnable projections of the vehicle state embeddings, $\mathbf{q}_i=\mathbf{W}^Q\mathbf{h}_{i,t}$, $\mathbf{k}_j=\mathbf{W}^K\mathbf{h}_{j,t}$ and $\mathbf{v}_j=\mathbf{W}^V\mathbf{h}_{j,t}$. The relevance $\omega_{ij}$ of vehicle $j$ to vehicle $i$ is
\begin{equation}
\omega_{ij}=\frac{\exp\!\big(\mathbf{q}_i^\top\mathbf{k}_j/\sqrt{d_k}+M_j\big)}{\sum_{l}\exp\!\big(\mathbf{q}_i^\top\mathbf{k}_l/\sqrt{d_k}+M_l\big)},
\label{eq:relevance}
\end{equation}
where $d_k$ is the key dimension. The embeddings $\mathbf{h}_{i,t}$ of all vehicles are stacked by vehicle ID into a fixed-size tensor, and slots without a vehicle are excluded with $M_j$, the vehicle padding mask, which is $0$ for occupied slots and $-\infty$ otherwise.

The output of a single attention head for vehicle $i$ is the relevance-weighted sum of the values,
\begin{equation}
\mathbf{u}_{i,t}=\sum_{j}\omega_{ij}\,\mathbf{v}_j.
\label{eq:attention}
\end{equation}

In addition, deciding a vehicle's manoeuvres requires taking into account multiple aspects of the traffic situation, such as avoiding collision, leaving a sufficient merge gap, and minimising the effect on following vehicles. Each aspect may require attention to different vehicles. For example, a vehicle may focus more on its neighbouring vehicles to avoid collisions, and may focus more on the vehicles near the merge point to yield a sufficient gap. We therefore use a multi-head attention architecture, in which each self-attention head has its own projections $\mathbf{W}^Q$, $\mathbf{W}^K$ and $\mathbf{W}^V$, allowing a vehicle to attend to different vehicles for several criteria simultaneously.

To accurately capture the relevance information in complex interactions between vehicles, we stack $L$ transformer encoder layers. The output of the final layer for vehicle $i$ at timestep $t$ is denoted by $\mathbf{z}_{i,t}$, which is then concatenated with vehicle $i$'s own embedding $\mathbf{h}_{i,t}$. The output combines the vehicle's own state with the states of its relevant vehicles

\paragraph{Actor-critic network.}
We train our traffic control model with the proximal policy optimisation (PPO) algorithm~\cite{ppo}, which requires an actor and a critic network. The actor is the control policy, which maps the state of a vehicle to an action distribution. The critic estimates the value of the state of the vehicle, which helps the training. Both networks take the concatenated $[\mathbf{z}_{i,t}, \mathbf{h}_{i,t}]$ from the vehicle transformer encoder as input.

\subsection{Learning Algorithm}

We train the traffic control neural network with the standard PPO~\cite{ppo} algorithm, with some modifications for the multi-vehicle setting, where a single joint policy outputs the joint action that controls all vehicles.

\paragraph{Joint policy.}
Our traffic control network outputs an action distribution $\pi_\theta(\mathbf{a}_{i,t} \mid \mathbf{o}_t)$ for every vehicle in a single forward pass, and the joint policy is the product of the per-vehicle distributions,
\begin{equation}
\pi_\theta(\mathbf{a}_t \mid \mathbf{o}_t) = \prod_{i \in \mathcal{V}_t} \pi_\theta(a_{i,t} \mid \mathbf{o}_t).
\label{eq:factorised_policy}
\end{equation}
Each vehicle's distribution is conditioned on the joint observation $\mathbf{o}_t$ of all vehicles, and the vehicle transformer layer is trained to filter $\mathbf{o}_t$ for each vehicle.

\paragraph{Training objective.}
The traffic control actor network is trained with the joint PPO loss function $\mathcal{L}^{\mathrm{actor}}$, which averages over the performance of all vehicles:
\begin{equation}
\begin{aligned}
\mathcal{L}^{\mathrm{actor}}(\theta) &= -\frac{1}{N}\sum_{t}\sum_{i \in \mathcal{V}_t} \min\big(\rho_{i,t}\hat{A}_{i,t},\ \bar{\rho}_{i,t}\hat{A}_{i,t}\big),\\
\rho_{i,t} &= \frac{\pi_\theta(a_{i,t} \mid \mathbf{o}_t)}{\pi_{\theta_{\mathrm{old}}}(a_{i,t} \mid \mathbf{o}_t)},\\
\bar{\rho}_{i,t} &= \mathrm{clip}\big(\rho_{i,t}, 1-\epsilon, 1+\epsilon\big),
\end{aligned}
\label{eq:ppo_loss}
\end{equation}
where $\mathcal{V}_t$ is the set of vehicles in the control region at step $t$, $N = \sum_{t}|\mathcal{V}_t|$ is the number of vehicle steps, $\epsilon$ is the clipping range, and $\hat{A}_{i,t}$ is the advantage of the vehicle $i$.

During training, a maximum number of vehicles is assumed. When a vehicle enters the control region, an ID $i$ from a fixed number of slots is assigned to it. The vehicle's ID is released when it exits the control region and will be reused by a newly entering vehicle. The advantage $\hat{A}_{i,t}$ of vehicle $i$ is computed with generalised advantage estimation (GAE)~\cite{gae}, which indicates vehicle $i$'s performance relative to the value predicted by the critic network:
\begin{equation}
\begin{aligned}
\hat{A}_{i,t} &= m_{i,t}\big(\delta_{i,t} + \gamma\lambda\, m_{i,t+1}\, \hat{A}_{i,t+1}\big),\\
\delta_{i,t} &= r_{i,t} + \gamma\, m_{i,t+1}\, V_\phi(\mathbf{o}_{t+1})_i - V_\phi(\mathbf{o}_t)_i,
\end{aligned}
\label{eq:gae}
\end{equation}
where $r_{i,t}$ is the reward of vehicle $i$, $V_\phi(\mathbf{o}_t)_i$ is the value of vehicle $i$ predicted by the critic, $\gamma$ is the reward discount factor, $\lambda$ is the trace-decay parameter of GAE, and $m_{i,t}\in\{0,1\}$ indicates whether ID $i$ is occupied at step $t$. The mask $m_{i,t}$ resets the GAE of the vehicle with ID $i$ when it exits the control region, so that when a new vehicle later reuses the same ID $i$, the GAE of the previous vehicle will not be bootstrapped from the new vehicle.

The critic network is trained by minimising the mean squared error between its predicted values and the discounted returns:
\begin{equation}
\begin{aligned}
\mathcal{L}^{\mathrm{critic}}(\phi) &= \frac{1}{N}\sum_{t}\sum_{i \in \mathcal{V}_t} \big(V_\phi(\mathbf{o}_t)_i - R_{i,t}\big)^2,\\
R_{i,t} &= m_{i,t}\big(r_{i,t} + \gamma\, m_{i,t+1}\, R_{i,t+1}\big),
\end{aligned}
\label{eq:critic_loss}
\end{equation}
where $R_{i,t}$ is the discounted return of vehicle $i$, which is reset by the mask $m_{i,t}$ when the vehicle exits the control region, similar to the GAE.

\section{Experiment}
\begin{figure*}[t]
\centering
\includegraphics[width=\textwidth]{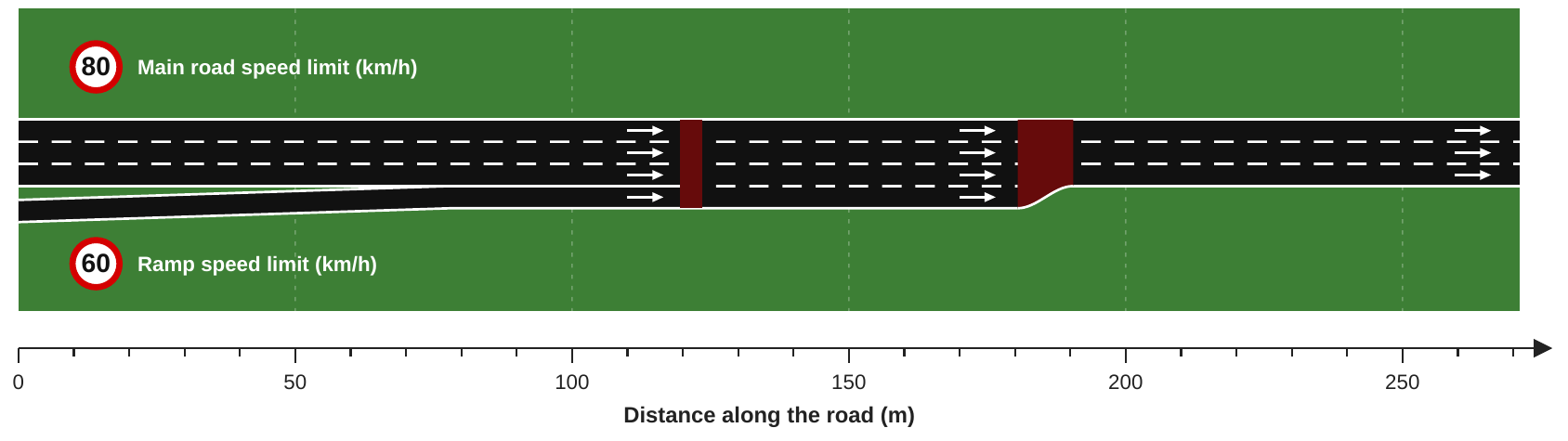}
\caption{Test scenario: a single-lane on-ramp joins the three-lane main road. Arrows indicate the driving direction.}
\label{fig:scenario}
\end{figure*}

We evaluate the proposed transformer-based traffic control policy (TRAN) in the SUMO simulator~\cite{sumo} against three baselines: the model (HOMO) in which each vehicle observes only its surroundings\cite{CTDE-baseline}, the CAE model (CAE) in which each vehicle observes the entire traffic\cite{CAE}, and the rule-based SUMO model (SUMO). The test scenario is shown in Fig.~\ref{fig:scenario}: A single-lane on-ramp joins the three-lane main road, and the control region is 300 m long. The speed limits are 80 km/h on the main road and 60 km/h on the ramp road, and the traffic flow is approximately 2400 veh/h/lane. 

We evaluate the performance of our model from four aspects: learning performance, which compares the training performance and the run-time performance of the models; long-run performance, which tests the models' robustness under changing traffic demand; performance with HDVs, which tests our model in mixed traffic with HDVs; and performance under communication errors and sensor noise. Besides the evaluation, we visualise the attention weights of TRAN to analyse its decisions, showing which vehicles an ego vehicle focuses on in different traffic situations.

\subsection{Model Training and Learning Performance}
\label{subsec:learning_performance}

We train TRAN, HOMO and CAE with PPO under the same settings for 300 episodes. Each episode simulates 5 minutes of traffic, and the vehicles are controlled every 100 ms. The SUMO model requires no training. The parameters of the reward function and the training are listed in Table~\ref{tab:constants}.

\begin{table}[t]
\centering
\begin{tabular}{llc}
\hline
Symbol & Description & Value \\
\hline
\multicolumn{3}{l}{\textit{Environment}} \\
 & Control section length & 300 m \\
 & Episode length & 5 min \\
 & Control interval & 100 ms \\
 & Lane-change cooldown time & 5 s \\
 & Number of vehicle IDs & 100 \\
\hline
\multicolumn{3}{l}{\textit{Reward}} \\
$\alpha$ & Weight of the global reward & 0.8 \\
$\beta$ & local reward offset & 0.2 \\
$\kappa_0$ & Base collision penalty & 50 \\
$\kappa_s$ & Collision severity scale & 5 \\
$\lambda_{\mathrm{tp}}$ & Throughput scale & 1/40 \\
$\lambda_{\mathrm{eb}}$ & Hard brake scale & 0.5 \\
\hline
\multicolumn{3}{l}{\textit{Network}} \\
$T$ & Observation history length & 5 s \\
$L$ & Number of transformer layers & 3 \\
$H$ & Number of attention heads & 4 \\
\hline
\multicolumn{3}{l}{\textit{Training}} \\
$\gamma$ & Discount factor & 0.98 \\
$\lambda$ & GAE parameter & 0.98 \\
 & Learning rate & $1\times10^{-4}$ \\
$\epsilon$ & PPO clip ratio & 0.15 \\
 & Entropy coefficient & 0.001 \\
 & Batch size & 50 \\
 & Epochs per update & 3 \\
 & Training episodes & 300 \\
\hline
\end{tabular}
\caption{Environment, reward, network and training parameters.}
\label{tab:constants}
\end{table}

\begin{figure*}[t]
    \centering
    \begin{subfigure}[t]{0.32\textwidth}
        \centering
        \includegraphics[width=\linewidth]{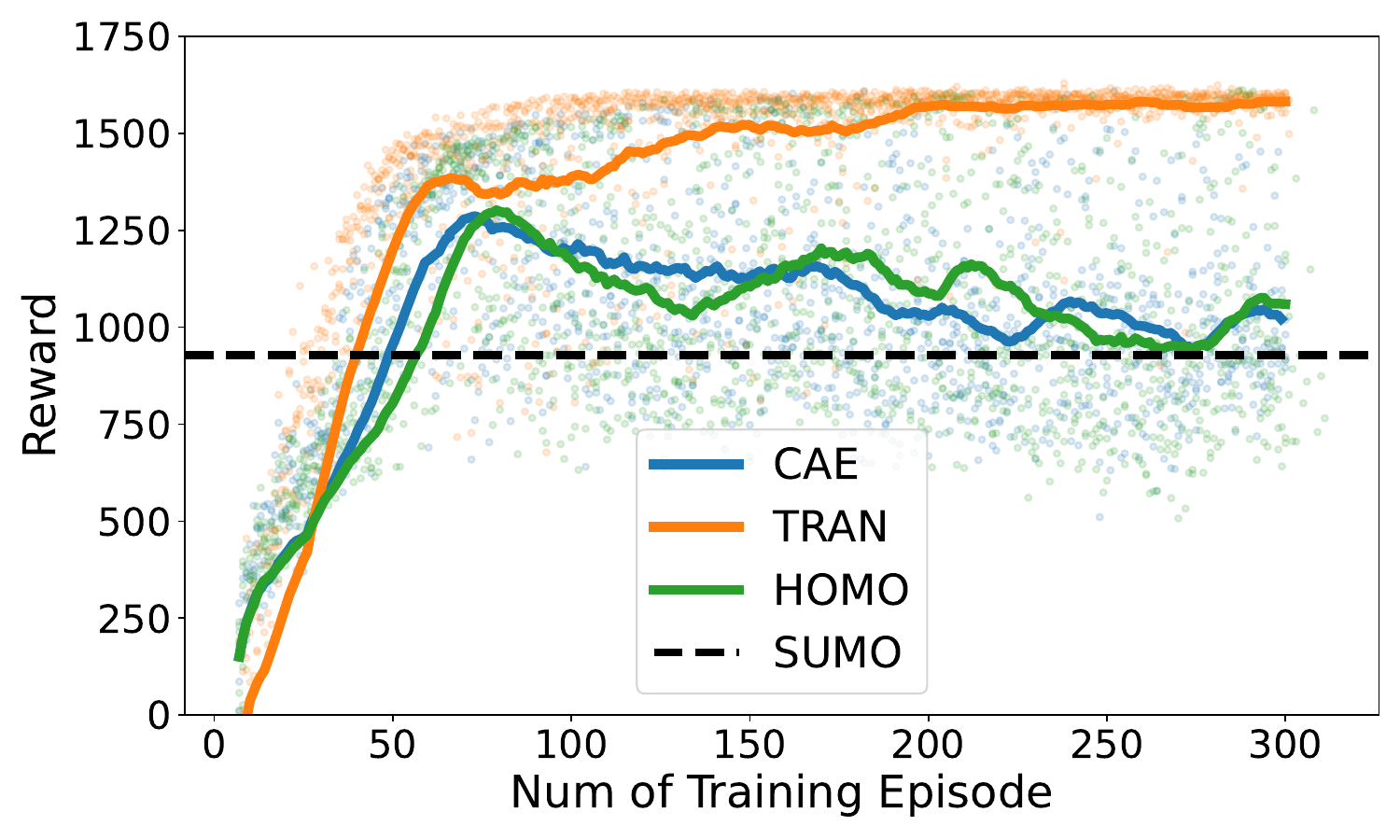}
        \caption{Episodic return}
        \label{fig:learning_reward}
    \end{subfigure}
    \hfill
    \begin{subfigure}[t]{0.32\textwidth}
        \centering
        \includegraphics[width=\linewidth]{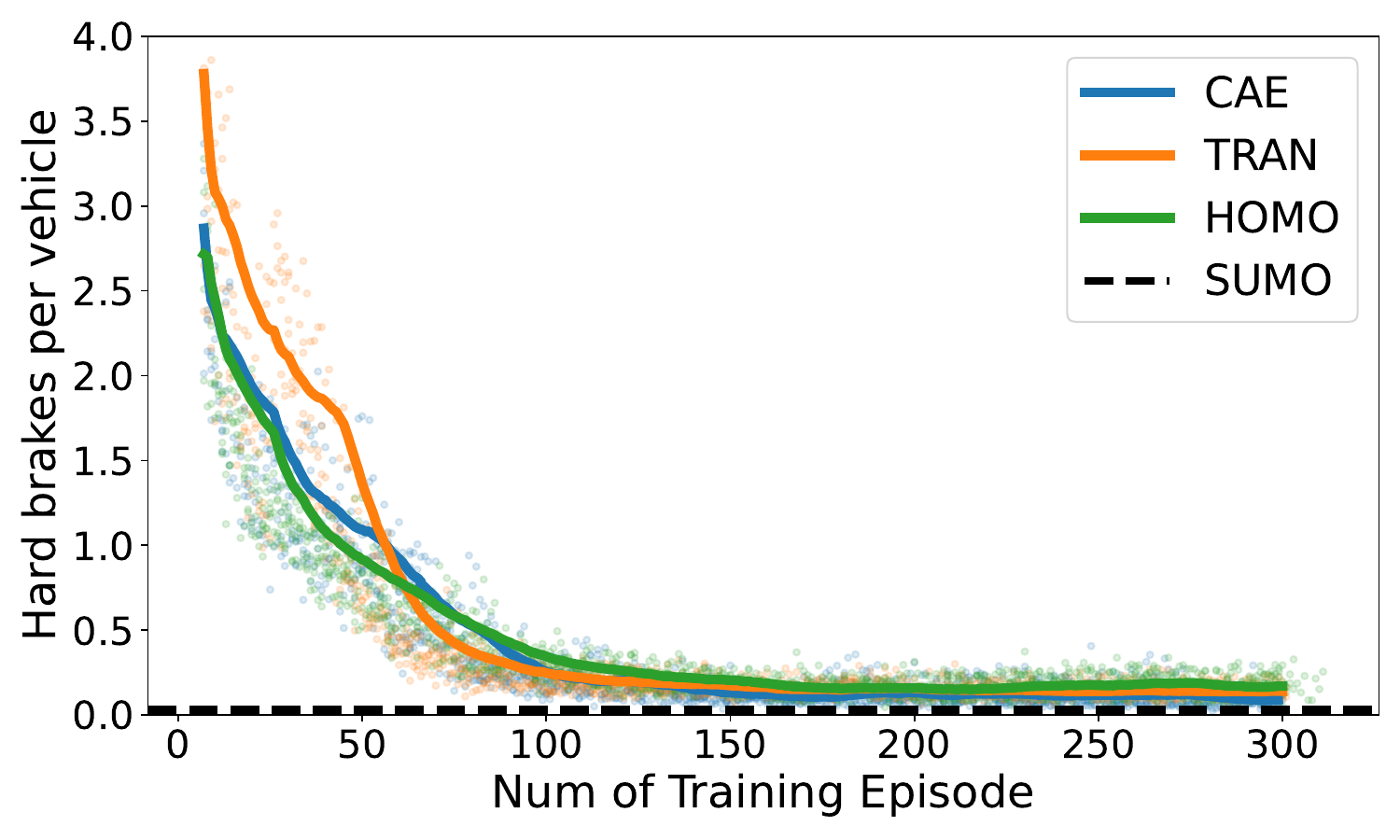}
        \caption{Number of hard brakes}
        \label{fig:learning_eb}
    \end{subfigure}
    \hfill
    \begin{subfigure}[t]{0.32\textwidth}
        \centering
        \includegraphics[width=\linewidth]{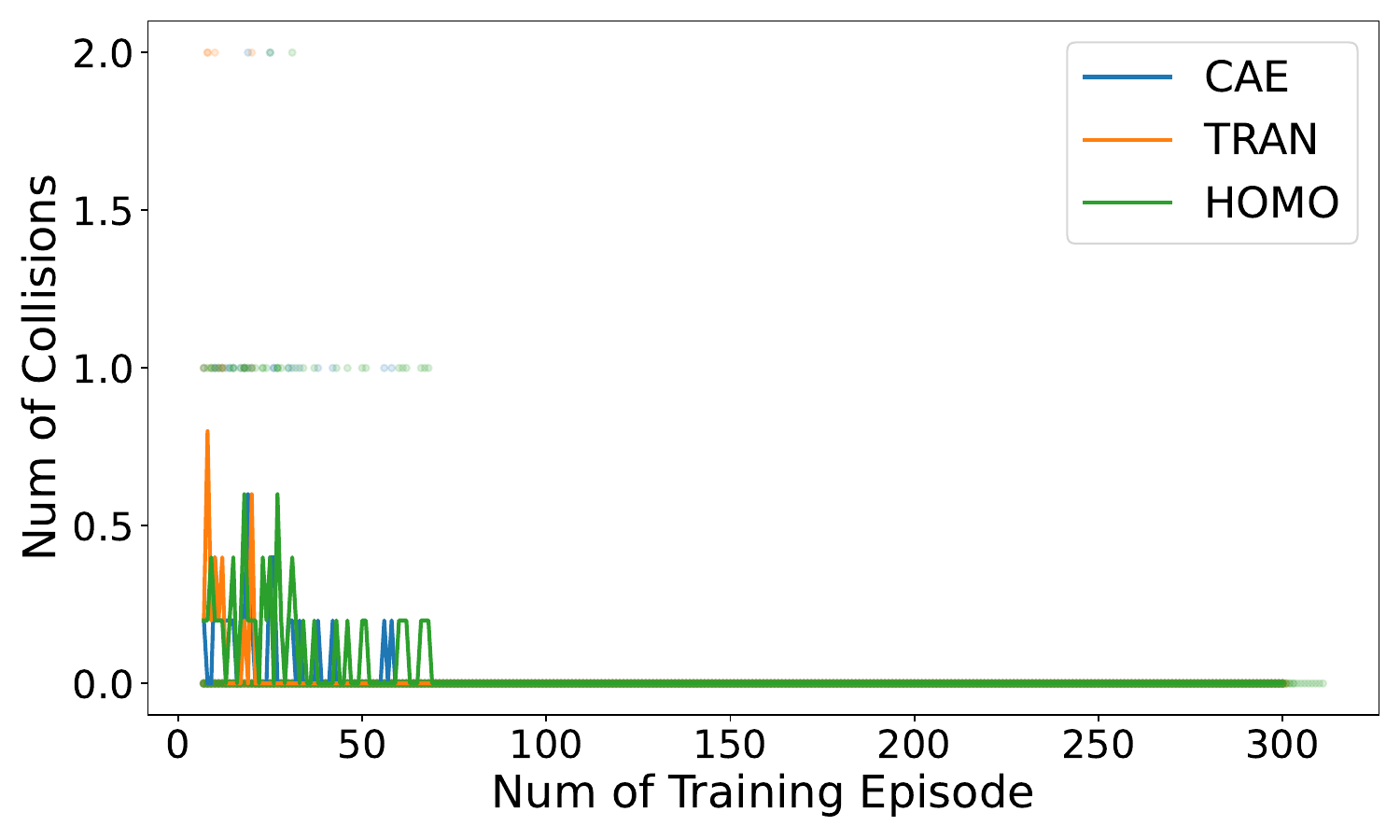}
        \caption{Number of collisions}
        \label{fig:learning_collision}
    \end{subfigure}
    \caption{Training performance of the compared models.}
    \label{fig:learning}
\end{figure*}

\begin{table*}[t]
    \centering
    \small
    \setlength{\tabcolsep}{5pt}
    \resizebox{\textwidth}{!}{%
    \begin{tabular}{lccccc}
        \toprule
        Model & \makecell{Total\\Reward} & \makecell{Hard Brakes\\/ Veh.} & \makecell{Main\\TF} & \makecell{Merge\\TF} & \makecell{Lane Changes\\/ Veh.} \\
        \midrule
        TRAN & $\mathbf{1604.75 \pm 11.83}$ & $0.106 \pm 0.025$ & $\mathbf{579.44 \pm 1.30}$ & $\mathbf{158.12 \pm 5.20}$ & $\mathbf{0.206 \pm 0.019}$ \\
        CAE  & $1001.83 \pm 128.26$ & $0.094 \pm 0.034$ & $485.04 \pm 18.06$ & $74.24 \pm 14.90$  & $0.348 \pm 0.027$ \\
        HOMO & $1339.81 \pm 231.11$ & $0.112 \pm 0.045$ & $532.52 \pm 37.04$ & $121.40 \pm 29.29$ & $0.249 \pm 0.026$ \\
        SUMO & $928.22 \pm 88.30$   & $0.032 \pm 0.015$ & $471.68 \pm 13.35$ & $86.96 \pm 7.70$   & $0.149 \pm 0.033$ \\
        \bottomrule
    \end{tabular}%
    }
\caption{Performance comparison of the models (mean $\pm$ standard deviation over 50 episodes). TF: traffic flow (veh per episode). Hard brakes and lane changes are averaged per vehicle.}
\label{tab:metrics-comparison}
\end{table*}

Fig.~\ref{fig:learning} shows the training performance of the models, averaged over five training runs. TRAN reaches a higher return than HOMO and CAE, which stays stable at around 1600 after 200 episodes. In contrast, the returns of HOMO and CAE peak after about 80 episodes and then drop to around 1000. However, the number of hard brakes keeps decreasing for all models (Fig.~\ref{fig:learning_eb}). This suggests that HOMO and CAE have difficulty reducing hard brakes while maintaining a high traffic flow. In other words, they cannot balance local and global collaboration; e.g., the vehicles can only slow down to reduce the interaction intensity and thereby avoid hard braking. Nevertheless, the returns of all learning-based models are higher than those of SUMO.

In terms of safety, collisions occur in all models at the beginning of training, but none occur after about 70 episodes (Fig.~\ref{fig:learning_collision}). The average number of hard brakes per vehicle also decreases for all models, but does not reach zero for the learning-based model or the rule-based SUMO model. Possible reasons are the intense interaction between vehicles, traffic uncertainties, or the stochasticity of the RL policy. In future work, we would like to explore ways to further reduce vehicle braking for smoother and more comfortable vehicle manoeuvres.

We further evaluate the performance of each method over 50 episodes, as listed in Table~\ref{tab:metrics-comparison}. In addition to the metrics in the reward function, we also report the average number of lane changes per vehicle, which is not included in the reward function. TRAN achieves the highest return and traffic flow. Moreover, TRAN has the fewest lane changes per vehicle among the learning-based models. This suggests that TRAN achieves a high traffic flow through better coordinated control of the vehicles. Although SUMO causes fewer hard brakes and lane changes, it results in a much lower traffic flow. 

Overall, our proposed transformer-based model achieves the best performance in both local and global collaboration: while achieving a higher and more stable return and traffic flow, it keeps hard brakes at a similar level to HOMO and CAE and requires fewer lane changes per vehicle. This indicates the effective design of the multi-head attention mechanism, which effectively filters the observations of relevant vehicles for each vehicle and allows each vehicle to focus on different aspects of the traffic situation. Besides the model architecture, the results also show the effectiveness of the reward function: all learning-based models become collision-free and reduce hard brakes during training, which indicates that the reward function guides the policies towards safe and collaborative behaviour.

\subsection{Long-Run Performance}
\label{subsec:long_run_performance}
\begin{figure}[t]
    \centering
    \begin{subfigure}[t]{\columnwidth}
        \centering
        \includegraphics[width=\linewidth]{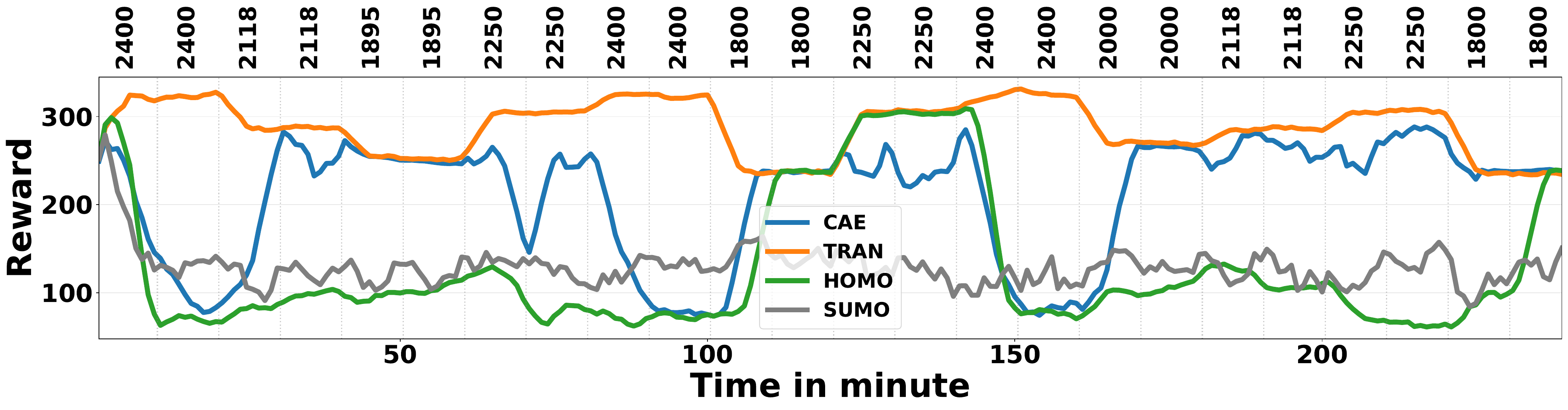}
        \caption{Reward}
        \label{fig:long_run_reward}
    \end{subfigure}
    \begin{subfigure}[t]{\columnwidth}
        \centering
        \includegraphics[width=\linewidth]{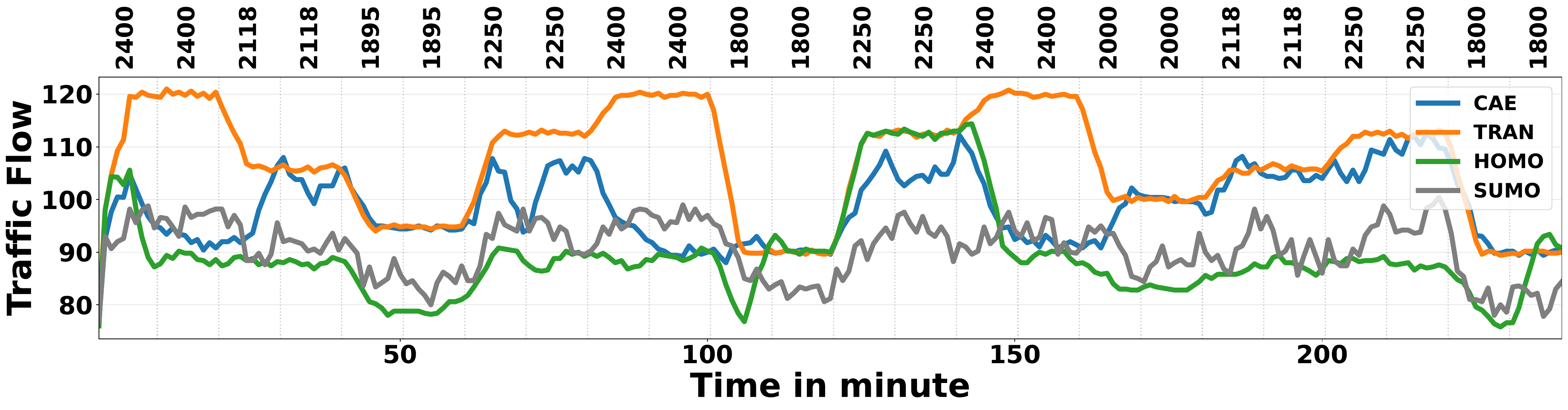}
        \caption{Main road traffic flow}
        \label{fig:long_run_main_flow}
    \end{subfigure}
    \begin{subfigure}[t]{\columnwidth}
        \centering
        \includegraphics[width=\linewidth]{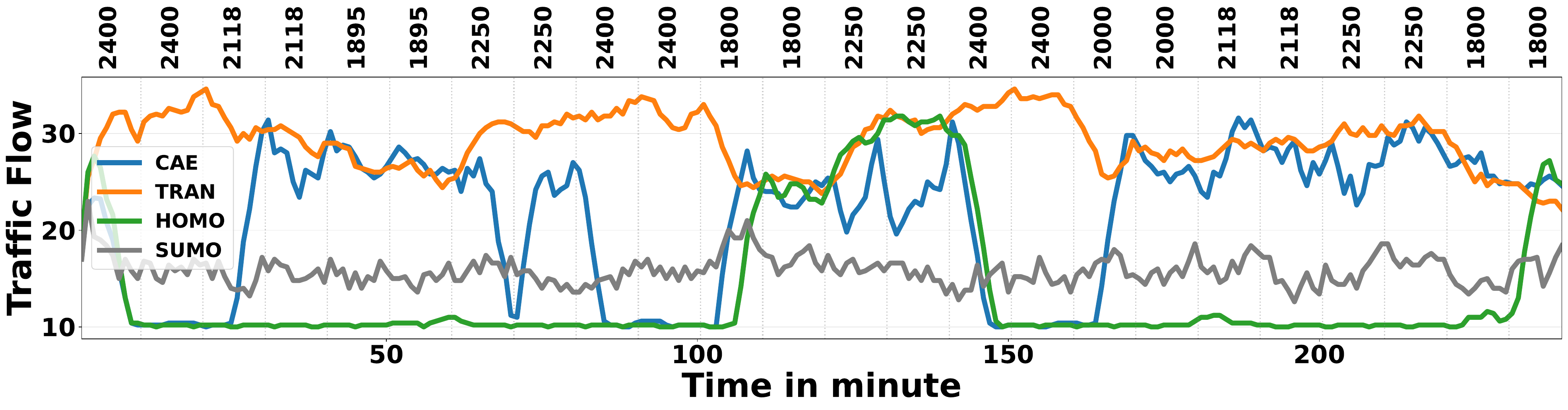}
        \caption{Merge ramp traffic flow}
        \label{fig:long_run_ramp_flow}
    \end{subfigure}
    \caption{Long-run performance under a four-hour random traffic flow. The labels at the top show the traffic demand in veh/h/lane. Traffic flow is the number of vehicles per minute.}
    \label{fig:long_run}
\end{figure}

To evaluate the robustness of the models under dynamic traffic demand, we run each model for a four-hour simulation in SUMO, during which the traffic flow starts at 2400 veh/h/lane and changes randomly every 10 minutes between 1800 and 2400 veh/h/lane. Fig.~\ref{fig:long_run} shows the reward and the traffic flows of the main road and the merge ramp.

Fig.~\ref{fig:long_run_reward} shows that TRAN keeps a high reward throughout the simulation. The resulting main road and merge ramp traffic flows follow the traffic demand (Figs.~\ref{fig:long_run_main_flow} and \ref{fig:long_run_ramp_flow}), which indicates the robustness of the transformer-based model. In contrast, CAE and HOMO fall into a traffic jam within the first 10 minutes, when the traffic flow is 2400 veh/h/lane. CAE recovers from the traffic jam when the traffic flow drops to about 2100 veh/h/lane, and then performs similarly to TRAN. However, its traffic becomes unstable at 2250 veh/h/lane, and it falls into a jam again if the traffic flow reaches 2400 veh/h/lane. HOMO has much worse performance. It can only recover from a traffic jam when the traffic flow drops to 1800 veh/h/lane, and it falls into a jam again when the traffic flow rises to 2400 veh/h/lane. In addition, when a traffic jam occurs, the performance of CAE and HOMO can be lower than that of SUMO.

These results show how the observation of each model affects its performance. HOMO only observes the surrounding vehicles, which is not sufficient for a vehicle to anticipate the traffic in regions beyond its neighbourhood, such as the merge point. CAE observes the global traffic state and therefore performs better than HOMO. However, CAE encodes the global traffic state with convolutional layers, which aggregate the information of nearby vehicles into region-level features. As a result, the individual state of each vehicle is represented coarsely, which makes local collaboration difficult and potentially causes the unstable performance of CAE.

\subsection{Performance with Human-Driven Vehicles}
\label{subsec:hdv_performance}

\begin{figure*}[t]
    \centering
    \begin{subfigure}[t]{0.32\textwidth}
        \centering
        \includegraphics[width=\linewidth]{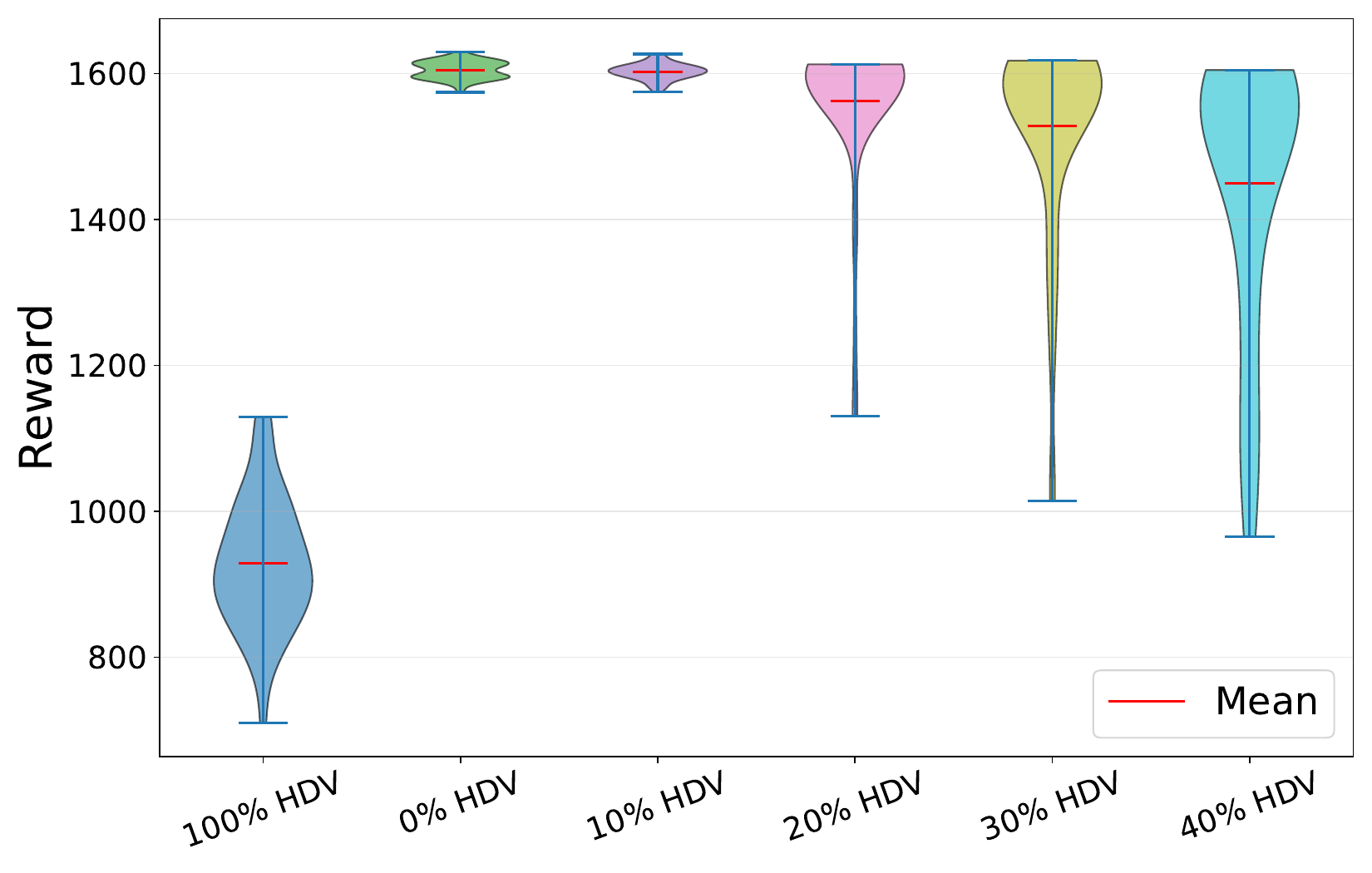}
        \caption{Reward}
        \label{fig:hdv_reward}
    \end{subfigure}
    \hfill
    \begin{subfigure}[t]{0.32\textwidth}
        \centering
        \includegraphics[width=\linewidth]{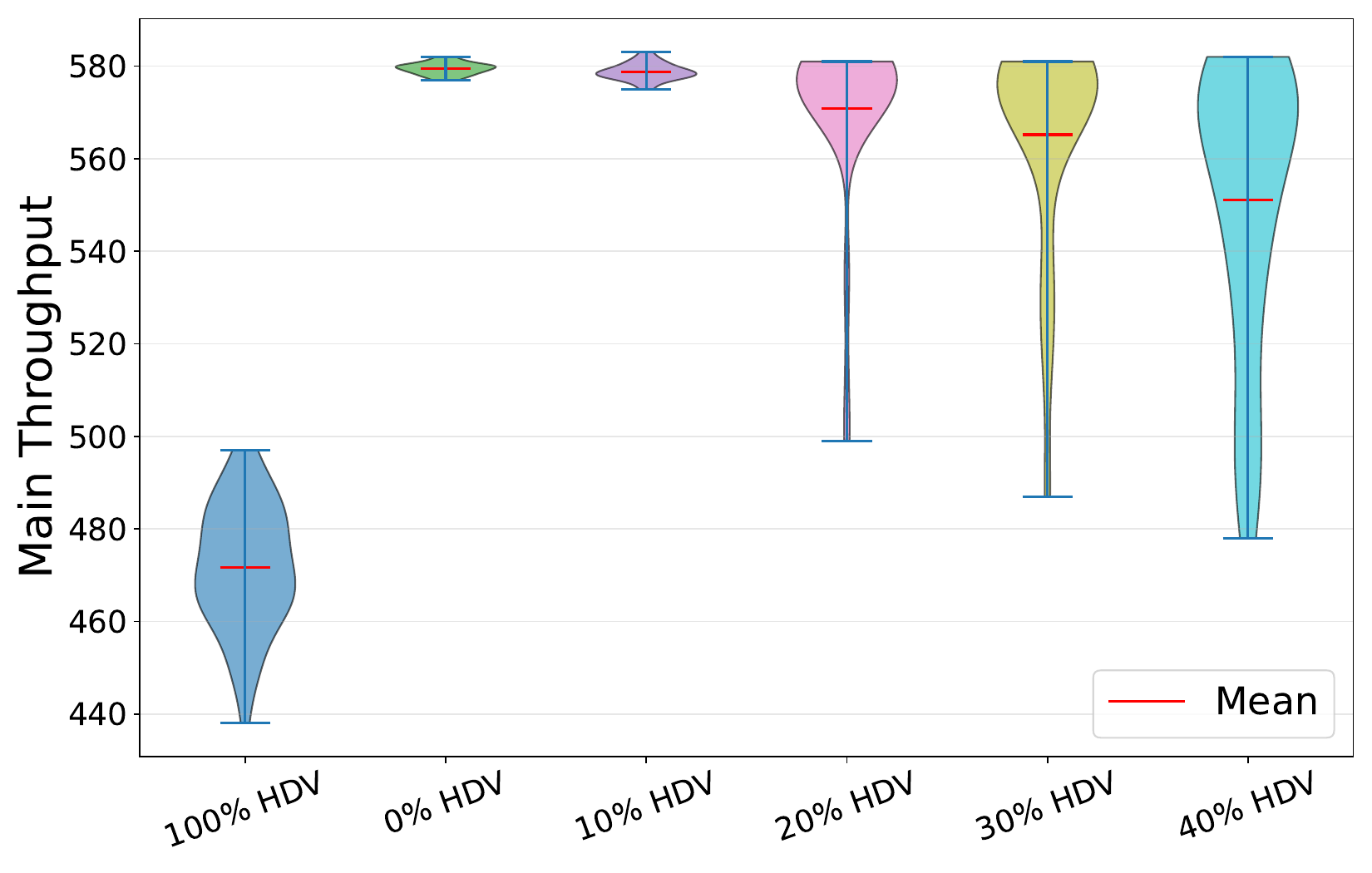}
        \caption{Main road traffic flow}
        \label{fig:hdv_main}
    \end{subfigure}
    \hfill
    \begin{subfigure}[t]{0.32\textwidth}
        \centering
        \includegraphics[width=\linewidth]{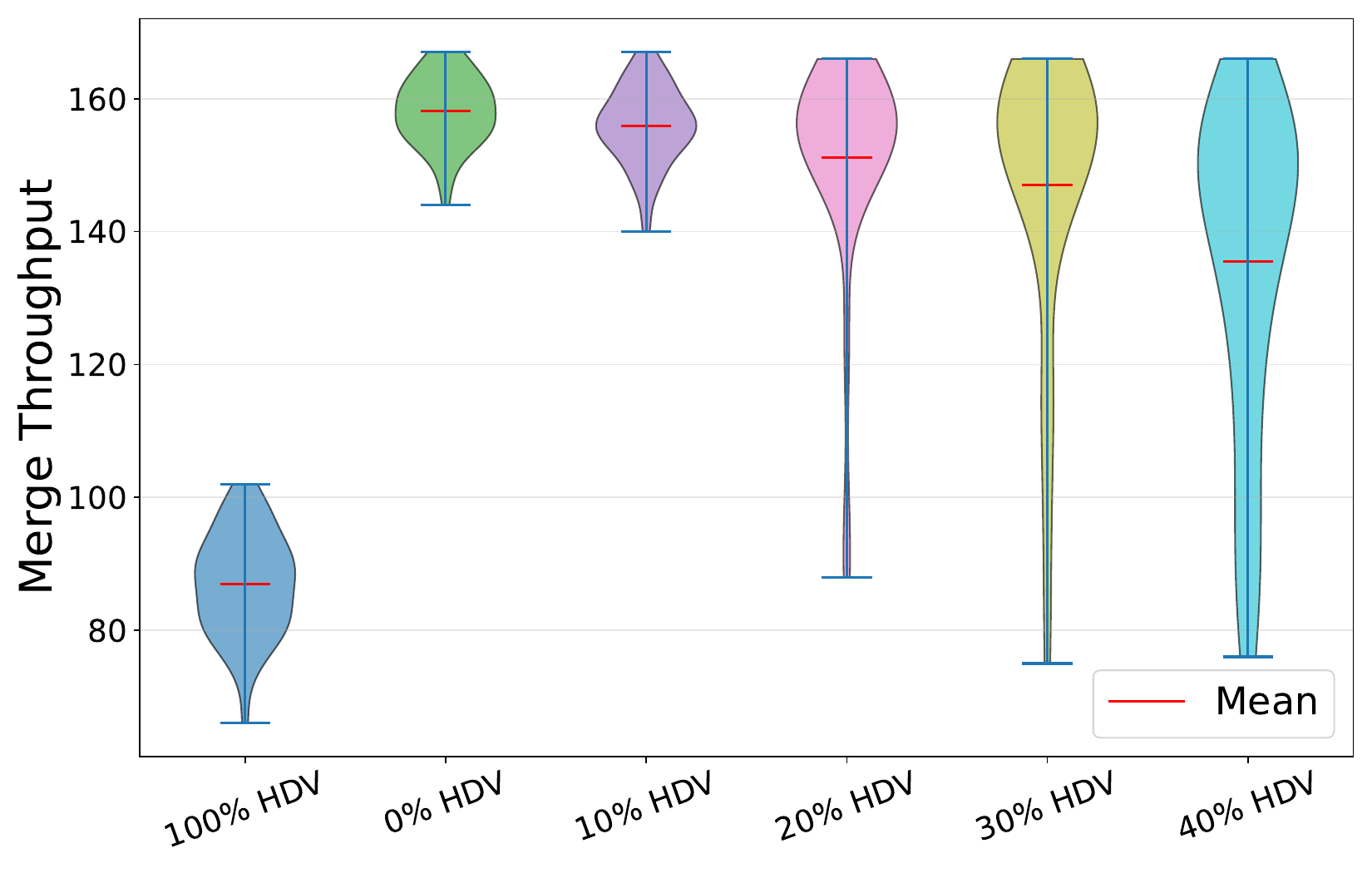}
        \caption{Merge ramp traffic flow}
        \label{fig:hdv_ramp}
    \end{subfigure}
    \caption{Performance of our model at different HDV shares. ``100\% HDV'' is the SUMO baseline.}
    \label{fig:hdv}
\end{figure*}

In this section, we evaluate the performance of TRAN in a mixed traffic scenario closer to real-world traffic, where CAVs share the road with human-driven vehicles (HDVs). We test TRAN with 10\% to 40\% HDVs and at a traffic flow of 2400 veh/h/lane. The HDVs are controlled by the vehicle control models of SUMO to simulate human driving behaviour, and they do not follow the actions of the policy. Each setting is evaluated over 50 episodes, and the results are shown in Fig.~\ref{fig:hdv}.

With 10\% HDVs, the performance of TRAN is almost the same as without HDVs. From 20\% HDVs, the mean reward and traffic flows decrease slightly, but their variance increases, because in some episodes the uncoordinated HDVs can cause congestion at the merge point. As the share of HDVs increases, the congestion becomes more frequent and more severe. At 40\% HDVs, the performance in the worst episodes can fall to the level of SUMO.

Nevertheless, TRAN outperforms SUMO in the mean reward and traffic flows at all HDV shares. This shows that TRAN can still improve the traffic when only part of the vehicles is controllable, but the improvement decreases as the share of HDVs increases. The results also show that a small group of uncoordinated vehicles can make the traffic unstable, possibly because a single HDV that changes lanes without considering other vehicles can block the merge and cause a jam that the controlled vehicles cannot easily resolve in dense traffic. In future work, we would like to explore how CAVs can better anticipate and adapt to the behaviour of HDVs in mixed traffic.

\subsection{Robustness to Communication Constraints and Sensor Noise} 
\label{sec:robustness} 

The previous experiments evaluate our proposed model under ideal conditions, where all vehicles' sensors have no noise and vehicles can communicate with the infrastructure controller instantly without communication errors. In this experiment, we evaluate the performance of the model in the presence of the following real-world communication constraints and sensor noise:

\paragraph{Communication constraints} We assume each vehicle transmits its local state to the infrastructure controller at \SI{10}{\hertz}. As vehicles enter the control region at different times, we assume asynchronous communication between vehicles and the infrastructure-based controller, i.e., every vehicle has a different transmission phase. In addition, we consider message loss and delay in both the uplink (from the vehicles to the controller) and the downlink (from the controller to the vehicles).

We model the message loss using the Gilbert--Elliott model~\cite{Gilbert, Elliott}, a two-state Markov chain that generates bursts of consecutive message losses commonly observed in wireless channels. Message delays are sampled from a truncated Gaussian distribution with jitter (the standard deviation of the delay), which may cause out-of-order messages. The vehicle states manager in the controller therefore retains only the most recent message from each vehicle and discards older arrivals. If the controller has not received a message from a vehicle within a specified timeout period, the vehicle is considered unreachable and will be removed from the controller. 

Commands sent from the controller to the vehicles are subject to the same communication constraints. However, we assume roadside infrastructure has more capable communication hardware than on-board units. Therefore, we assume the downlink message loss and delay to be half of the uplink values.

We test the vehicle performance with three communication conditions: mild, normal, and stress. The mean message loss ranges from \SI{1}{\percent} in the mild condition, matching the near-zero packet error ratio measured in line-of-sight V2X communication experiments~\cite{dlr142341}, to \SI{30}{\percent} in the stress condition, the highest packet error ratio considered by SAE J2945/1 standard~\cite{sae_j2945_1}. The mean delay ranges from \SI{20}{\milli\second} in the mild condition, matching the latencies measured on real-world ITS-G5 and C-V2X links~\cite{dlr142341}, to \SI{100}{\milli\second} in the stress condition, the maximum latency allowed by the 3GPP standard for V2I communication~\cite{3gpp}. The normal condition lies at the midpoint between the two. We set the jitter to half of the mean delay and the mean burst length of lost messages to 2, 3.5 and 5 messages in the mild, normal and stress conditions.

\paragraph{Sensor noise} Vehicle positions are assumed to be measured by GNSS, with a time-correlated error and a constant vehicle-specific bias. Speed, acceleration, and heading are measured with independent Gaussian noise. In all conditions, the noise magnitudes match the minimum accuracy required by SAE J2945/1 standard~\cite{sae_j2945_1} for V2X on-board systems (68\% of measurements): \SI{1.5}{\metre} for position, \SI{0.28}{\metre\per\second} for speed, \SI{2}{\degree} for heading above \SI{12.5}{\metre\per\second} (\SI{3}{\degree} below), and \SI{0.3}{\metre\per\second\squared} for acceleration.

\begin{table}[t]
\centering
\resizebox{\columnwidth}{!}{%
\begin{tabular}{lccc}
\toprule
Parameter & Mild & Normal & Stress \\
\midrule
\multicolumn{4}{l}{\textit{Uplink}} \\
Mean message loss (\%)      & 1   & 15  & 30  \\
Mean burst length (msgs)    & 2.0 & 3.5 & 5.0 \\
Mean delay (ms)             & 20  & 60  & 100 \\
Jitter (ms)                 & 10  & 30  & 50  \\
\midrule
\multicolumn{4}{l}{\textit{Downlink}} \\
Mean message loss (\%)      & 0.5 & 7.5 & 15  \\
Mean delay (ms)             & 10  & 30  & 50  \\
Jitter (ms)                 & 5   & 15  & 25  \\
\midrule
\multicolumn{4}{l}{\textit{Sensor (all degrees)}} \\
GNSS position, 68\% (m)             & \multicolumn{3}{c}{1.5} \\
Speed (m/s)                         & \multicolumn{3}{c}{0.28} \\
Acceleration (m/s$^2$)              & \multicolumn{3}{c}{0.3} \\
Heading ($^\circ$), speed $> 12.5$ m/s & \multicolumn{3}{c}{2} \\
Heading ($^\circ$), speed $\le 12.5$ m/s & \multicolumn{3}{c}{3} \\
\bottomrule
\end{tabular}%
}
\caption{Parameters of the communication errors and sensor noise.}
\label{tab:robustness_params}
\end{table}

Table~\ref{tab:robustness_params} lists the parameters of the three conditions. 

\begin{table*}[t]
    \centering
    \small
    \setlength{\tabcolsep}{5pt}
    \begin{tabular}{lcccccc}
        \toprule
        Condition & Total Reward & Collision & Hard Brake / Veh. & Main TF & Merge TF & Lane Changes / Veh. \\
        \midrule
        Ideal  & $1604.75 \pm 11.83$ & 0 & $0.106 \pm 0.025$ & $579.44 \pm 1.30$ & $158.12 \pm 5.20$ & $0.206 \pm 0.019$ \\
        Mild   & $1578.78 \pm 10.79$ & 0 & $0.155 \pm 0.021$ & $579.00 \pm 2.12$ & $156.67 \pm 5.37$ & $0.212 \pm 0.017$ \\
        Normal & $1548.21 \pm 14.02$ & 0 & $0.226 \pm 0.043$ & $578.27 \pm 1.80$ & $157.50 \pm 5.06$ & $0.248 \pm 0.024$ \\
        Stress & $1542.58 \pm 15.28$ & 0 & $0.245 \pm 0.041$ & $578.70 \pm 1.49$ & $157.43 \pm 5.02$ & $0.242 \pm 0.018$ \\
        \bottomrule
    \end{tabular}
\caption{Performance of the proposed policy under ideal conditions and the three communication and sensor noise conditions (mean $\pm$ standard deviation over 50 episodes). TF: traffic flow (veh per episode).}
\label{tab:robustness_results}
\end{table*}

For each condition, we run the proposed model for 50 evaluation episodes without any further training. Table~\ref{tab:robustness_results} shows the experiment results. From the results, we can see that the proposed model still performs well under all three conditions. The total reward decreases only slightly even in the stress condition, and both the main road traffic flow and the merging lane traffic flow remain almost unchanged. In addition, no collision occurs in any of the conditions. However, the hard brake rate per vehicle increases noticeably, by about \SI{46}{\percent}, \SI{113}{\percent}, and \SI{131}{\percent} in the mild, normal, and stress conditions. In addition, the average number of lane changes per vehicle increases by about \SI{20}{\percent} in the normal and stress conditions. Both are potentially related to a misjudgment of the available gap, for example, caused by inaccurate vehicle states resulting from sensor errors or message loss. A vehicle may then believe that a gap is available and change lanes or merge into the gap that is actually too small, and the vehicle behind it has to brake hard.

In summary, the results show that the proposed model is robust in terms of traffic efficiency under various real-world communication constraints and sensing errors. However, its safety and traffic smoothness performance degrade as the conditions get worse. In future work, we aim to improve the safety awareness and traffic smoothness of the model under such conditions while maintaining the traffic efficiency.

\subsection{Attention Visualisation}

\begin{figure*}[p]
    \centering
    \begin{subfigure}[t]{\textwidth}
        \centering
        \includegraphics[width=0.95\linewidth]{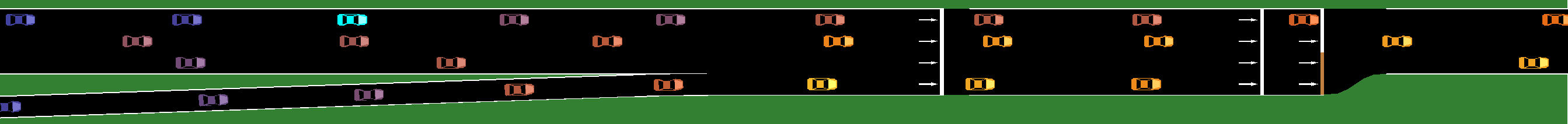}\\[0.5mm]
        \includegraphics[width=0.95\linewidth]{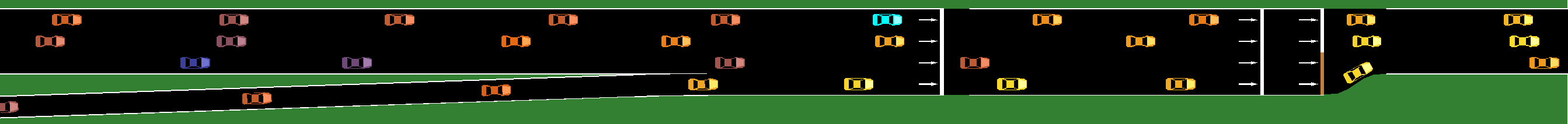}\\[0.5mm]
        \includegraphics[width=0.95\linewidth]{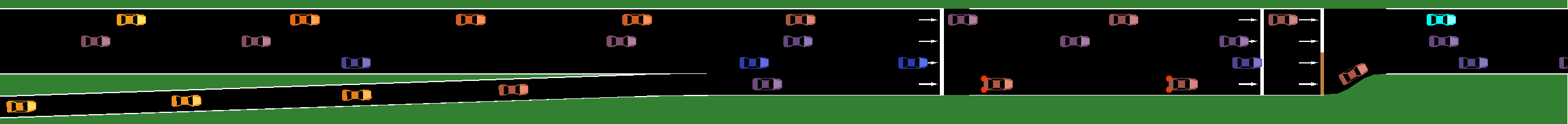}
        \caption{Left lane}
        \label{fig:attn_left}
    \end{subfigure}

    \vspace{1.5mm}
    \begin{subfigure}[t]{\textwidth}
        \centering
        \includegraphics[width=0.95\linewidth]{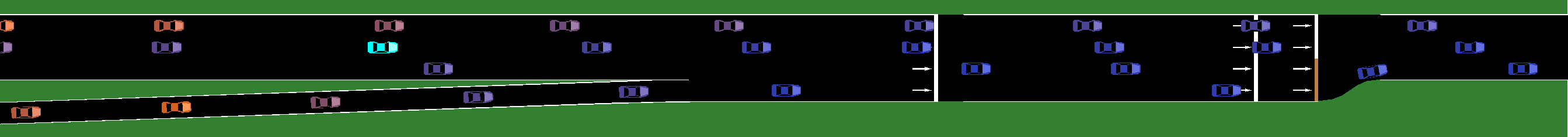}\\[0.5mm]
        \includegraphics[width=0.95\linewidth]{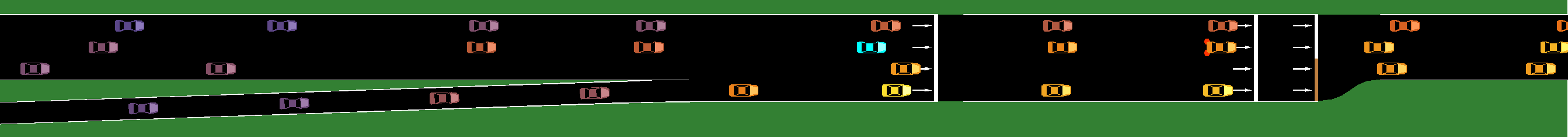}\\[0.5mm]
        \includegraphics[width=0.95\linewidth]{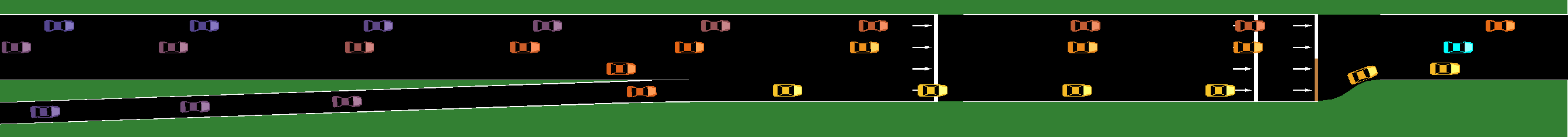}
        \caption{Middle lane}
        \label{fig:attn_middle}
    \end{subfigure}

    \vspace{1.5mm}
    \begin{subfigure}[t]{\textwidth}
        \centering
        \includegraphics[width=0.95\linewidth]{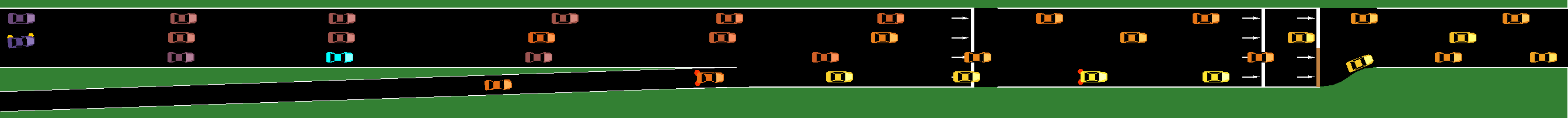}\\[0.5mm]
        \includegraphics[width=0.95\linewidth]{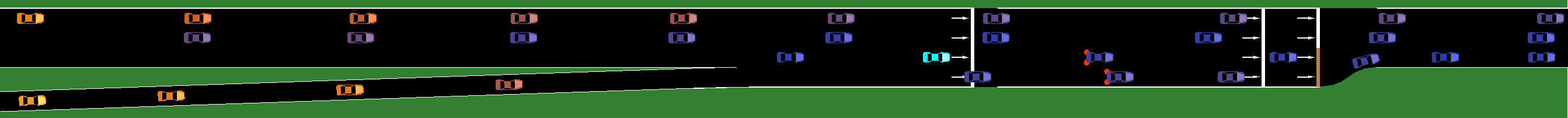}\\[0.5mm]
        \includegraphics[width=0.95\linewidth]{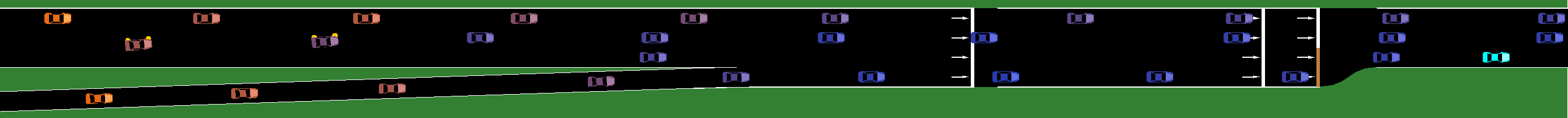}
        \caption{Right lane}
        \label{fig:attn_right}
    \end{subfigure}

    \vspace{1.5mm}
    \begin{subfigure}[t]{\textwidth}
        \centering
        \includegraphics[width=0.95\linewidth]{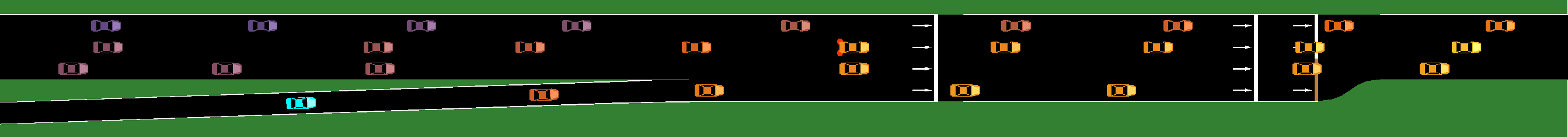}\\[0.5mm]
        \includegraphics[width=0.95\linewidth]{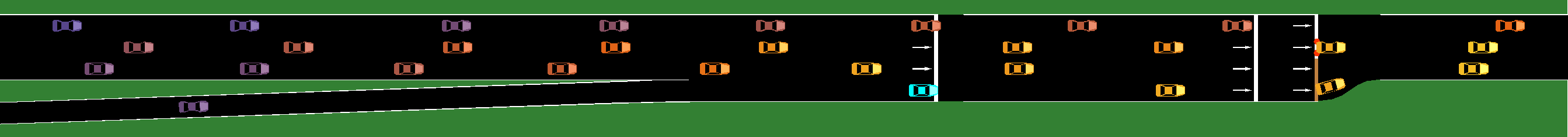}\\[0.5mm]
        \includegraphics[width=0.95\linewidth]{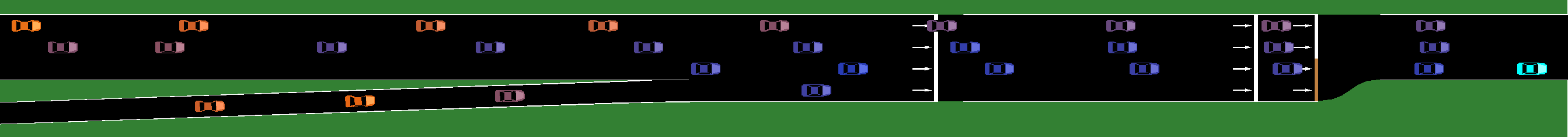}
        \caption{Merge Ramp}
        \label{fig:attn_ramp}
    \end{subfigure}
    \caption{Attention weights of an ego vehicle (cyan) starting in different lanes. In each subfigure, the three snapshots show the ego vehicle entering, inside and leaving the control region (top to bottom). Attention increases from dark blue (low) through orange (medium) to bright yellow (high).}
    \label{fig:attention}
\end{figure*}

In this section, we attempt to interpret the reasoning of the transformer-based vehicle control model by visualising how an ego vehicle assigns relevance to other vehicles in different situations. Fig.~\ref{fig:attention} shows four ego vehicles starting in each lane of the highway section. For each ego vehicle, three snapshots are taken corresponding to the times when it enters, is inside and leaves the control region. The other vehicles are coloured according to the attention weights assigned to them by the ego vehicle.

Although the attention distribution differs between situations, several common patterns can be observed. The transformer-based model does not assign a fixed relevance to the neighbouring vehicles or to all vehicles. Instead, the relevance of a vehicle changes with the traffic situation and the position of the ego vehicle. Before and inside the control region, the ego vehicle tends to focus on the vehicles around the merging point, where the most intensive interaction between vehicles takes place. After leaving the merging area, the ego vehicle shifts its attention to the vehicles entering the control region. A possible reason is that the ego vehicle no longer needs intensive interaction with its neighbours, while the vehicles behind it may still be affected by its actions. Together with the performance results, this confirms the effectiveness of our model design, which lets each vehicle focus only on the relevant vehicles.

However, not every attention distribution can be explained intuitively. For example, the ego vehicle in the middle lane attends to the vehicles upstream when entering the control region, unlike the ego vehicles in the other lanes, and it still attends to its neighbouring vehicles when leaving the control region. In addition, the attention weights only reflect the input that the model relies on, but they cannot fully explain the reasoning behind its decisions. Further studies are needed to investigate how the attention mechanism assigns weights to vehicles in different situations, and how the attended vehicles’ information is used for the ego vehicle’s final control decision.

\section{Conclusion}

In this paper, we proposed a transformer-based multi-agent reinforcement learning model (TRAN) for the cooperative on-ramp merging problem, where connected autonomous vehicles (CAVs) send their current state to a roadside infrastructure controller and receive control commands via vehicle-to-infrastructure (V2I) communication. To efficiently gather the information needed for vehicle control in dense traffic, each vehicle filters the relevant vehicles from the global traffic state using the proposed attention mechanism. The model is trained with a reward function that evaluates safety, traffic throughput and traffic smoothness.

We evaluated the proposed model's performance by comparing it with three baseline models: the model where vehicles only observe the neighbouring vehicles (HOMO), the CAE-based model where vehicles observe the entire traffic (CAE), and the rule-based SUMO model. The experiment results show that TRAN achieves the highest performance during training and evaluation. It results in the highest traffic flow on both the main road and the merge ramp, while requiring the fewest lane changes per vehicle among the learning-based models. All learning-based models achieve collision-free driving, and the number of hard brakes during training decreases throughout training. This shows that the reward function effectively guides the policies towards safe and smooth driving. We further test the models' performance in a four-hour simulation with dynamic traffic demand. TRAN never falls into a traffic jam, while HOMO and CAE fail at high traffic flow. TRAN can also improve the traffic performance in mixed traffic with human-driven vehicles (HDVs), although the performance becomes less stable when the HDVs' share increases. Moreover, TRAN remains robust in overall performance under communication errors and sensor noise, and no collision is observed, although its safety awareness degrades. We visualise the relevance weights that each vehicle assigns to other vehicles. The result shows that the attention assignments depend on the current traffic situation and the ego vehicle's road position, instead of a fixed neighbourhood observation or a uniform encoding of the entire traffic. This confirms the effectiveness of our design.

In future work, we would like to address the following directions:
\begin{itemize}
    \item improving the safety awareness of the model under poor communication and sensing conditions, to further reduce braking;
    \item enabling CAVs to better anticipate and adapt to the behaviour of HDVs in mixed traffic;
    \item investigating how the attention mechanism assigns weights to vehicles in different situations, and how the attended vehicles' information is used for the final control decision.
\end{itemize}

\clearpage
\bibliographystyle{plainnat}
\bibliography{reference.bib}

\end{document}